# Monitoring Horses in Stalls: From Object to Event Detection

Dmitrii Galimzianov, Viacheslav Vyshegorodtsev, Ivan Nezhivykh

*emble.pro*

## Abstract

Monitoring the behavior of stalled horses is essential for early detection of health and welfare issues but remains labor-intensive and time-consuming. In this study, we present a prototype vision-based monitoring system that automates the detection and tracking of horses and people inside stables using object detection and multi-object tracking techniques. The system leverages YOLOv11 and BoT-SORT for detection and tracking, while event states are inferred based on object trajectories and spatial relations within the stall. To support development, we constructed a custom dataset annotated with assistance from foundation models CLIP and GroundingDINO. The system distinguishes between five event types and accounts for the camera's blind spots. Qualitative evaluation demonstrated reliable performance for horse-related events, while highlighting limitations in detecting people due to data scarcity. This work provides a foundation for real-time behavioral monitoring in equine facilities, with implications for animal welfare and stable management.

## Keywords

Stalled horses, animal welfare, monitoring systems, event detection, object detection, object tracking, YOLO, SORT

## 1. Introduction

### 1.1. Problem Definition

Monitoring the daily behavior of horses plays a critical role in their care and management, facilitating the early identification of health problems and abnormal behaviors. However, such continuous observation is resource-intensive and time-consuming, often leading to inadequate individual monitoring [1].

To address this issue, numerous startups have developed various horse monitoring systems. These can be broadly categorized into two groups. The first group targets monitoring horse activity outdoors using specialized trackers[1,2,3,4,5,6]. The second group focuses on monitoring stalled (indoor) horses, typically employing Machine Learning (ML) and Computer Vision (CV) techniques[7,8,9,10]. Although the functionalities of the latter systems vary, the primary objective is the detecting and tracking horses within their stables [2]. However, the majority of these solutions do not disclose the underlying ML algorithms or their performance metrics.

---

[1] https://www.hoofstep.com/app-1
[2] https://www.equimetrics.ie/v-pro/
[3] https://horsano.com/en/
[4] https://www.steedems.com/
[5] https://ponyuptechnologies.com/
[6] https://horsepal.com/

### 1.2. Existing Work

Several research efforts have explored related areas. Delgado et al. compiled and annotated a dataset of 10,000 images of stalled horses in different postures and trained a YOLOv3 [3] model for object detection [2]. Although tracking was mentioned, specific methods for this task were not reported.

Kholiavchenko et al. developed a dataset for wildlife behavior recognition (e.g., zebras) using drone footage and implemented a pipeline combining YOLOv8 [4] and the SORT [5] tracking algorithm [6]. Building on this, Chan et al. introduced the YOLO-Behaviour framework for automatic detection of animal behavioral events [7]. An alternative approach was proposed by Kil et al., who employed the Loopy[11] horse pose estimation model based on anatomical landmarks to analyze the behavior of stabled horses [8].

Similar efforts have been made in cattle monitoring. Tassinari et al. utilized YOLOv3 for cow detection in free-stall barns [9], while Mon et al. proposed a custom tracking method based on bounding boxes from YOLOv8 [10].

### 1.3. Commonalities and Limitations

Three key observations emerge from these studies:

1. Despite differing objectives—ranging from wildlife behavior analysis to cattle identification—all tasks are ultimately framed as object detection and tracking problems. YOLO combined with SORT remains a popular and effective approach.
2. While many datasets were manually labeled, we found no work on leveraging powerful foundation models such as GroundingDINO [11] or CLIP [12] to streamline annotation.
3. The lack of standardized evaluation metrics hampers comparison across studies.

Most reported metrics assess performance on individual frames. For instance, Delgado et al. reported Intersection over Union (IoU) and mean Average Precision (mAP), while Tassinari et al. used an extended set of detection metrics. Kil et al. reported frame-based sensitivity, accuracy, and error rates.

Although object detection metrics are valuable intermediaries, they fail to capture the temporal dimension crucial in monitoring systems. Some authors attempted to address this. Mon et al. reported object tracking accuracy but did not define what constitutes a "correctly tracked" object [10]. Kholiavchenko et al. computed macro- and micro-averaged accuracy for action classification, but this ignored action time-bounds [6]. Chan et al. proposed event detection metrics, yet also neglected temporal localization [7].

---

[7] https://www.novostable.com/ailana
[8] https://www.horcery.com/stall-monitor-system
[9] https://eu.acaris.net/en/health-monitoring/
[10] https://developer.nvidia.com/blog/using-ai-to-monitor-horse-safety/
[11] http://loopb.io/

In real-time monitoring systems, both the classification and temporal localization of events are critical. The temporal Intersection over Union (t-IoU) metric [13] is a suitable measure for evaluating a system's ability to detect event start and end times.

### 1.4. Proposed Solution

Building upon the above studies, our solution utilizes YOLO and SORT as core components. The key contributions of our work are:

1. Employing CLIP and GroundingDINO to expedite dataset annotation.
2. Performing object tracking under complex conditions—distinguishing horses from people and handling blind spots (corners or edges of the stall that are not captured by the camera).
3. Designing an event detection system that identifies when horses or people are inside or outside a stall by estimating event start and end times.
4. Conducting qualitative analysis of the system's predictions to assess performance.

## 2. Materials and Methods

### 2.1. Horse Monitoring System

The overarching goal of our horse monitoring system is to detect and log various events occurring within a stall, such as abnormal horse behavior or interactions between horses and stablehands. As a foundational step toward this goal, we developed a prototype capable of detecting two key aspects:

1. Visibility – whether the horse is visible in the camera feed.
2. Location – whether the horse is inside its designated stall.

These two aspects yield four distinct events:

1. The horse is visible and inside the stall.
2. The horse is visible and outside the stall (i.e., in the adjacent hall).
3. The horse is invisible but inside the stall (i.e., in a blind spot).
4. The horse is invisible and outside the stall (i.e., has left the area).

A fifth event is defined when multiple horses are present in the stall simultaneously. The same five event types are applied analogously to people. To detect these events, we used an object detection neural network combined with a tracking algorithm as the core components of our system.

### 2.2. Dataset Collection

Each surveillance camera was positioned to cover a single stall, part of the adjacent hall, and occasionally a window. Night vision was activated during dark hours (see Figure 1). Since the system is intended for 24/7 real-time monitoring, it processes the input stream as one-minute video clips and returns predictions before the subsequent clip arrives. Videos are recorded at 20 frames per second with a resolution of 1280×720 pixels.

To train the object detection model, we collected and annotated a dataset of images containing horses and people. This was done in two iterations:

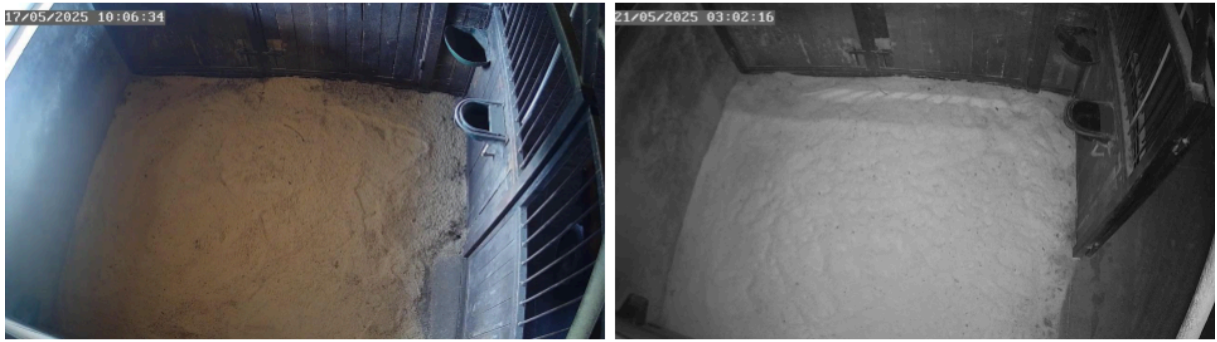

Figure 1. Empty stalls during the day and night time.

**First Iteration.** We selected one-minute clips from six different camera feeds. Given the time-consuming nature of manual labeling, we reduced the data volume through the following steps:

1. **Stratified Sampling:** We randomly selected clips, stratifying based on stall ID, time of day, and season.
2. **Frame Selection:** For each clip, we reduced the number of frames by taking every 60th frame. We then preserved only the most *informative* frames by computing cosine similarity between CLIP embeddings of consecutive frames (t and t–1) and retaining frames in the lowest 25th percentile of similarity—i.e., the most visually distinct.

We then performed automatic annotation using the GroundingDINO model, which was prompted to detect bounding boxes around horses and people. The automatically labeled images were imported into CVAT [14] for manual correction and refinement.

**Second Iteration.** We randomly selected clips from four of the six cameras used in the first round. This time, the automatic labeling was performed using a YOLOv11 [15] model trained on the previously labeled dataset. As before, all labels were refined manually in CVAT.

To assess the generalizability of our model, we held out one stall for validation. Table 1 shows the distribution of annotated objects (horses and people) across training and validation sets.

Table 1. Distribution of Objects Across Sets

| Stall | # Objects: Horses | # Objects: People | Split |
|---|---:|---:|---:|
| k1 | 2335 | 1749 | train |
| k2 | 4560 | 546 | train |
| k3 | 324 | 457 | validation |
| k4 | 262 | 221 | train |
| n1 | 1003 | 349 | train |
| n2 | 336 | 85 | train |

## 2.3. Multi-Object Detection and Tracking

We employed Ultralytics YOLOv11 for object detection. Multiple experiments were conducted to determine the optimal model size. Each model was trained for 25 epochs with a batch size of 16, using default training parameters on an NVIDIA Tesla T4 GPU.

Table 2 reports standard metrics (Precision, Recall, mAP50, and mAP50-95) on the validation set of the best checkpoints across experiments. The "M" model size was selected for further use based on the highest precision. For object tracking, we used the Ultralytics BoT-SORT algorithm with default parameters. During inference, two practical challenges were addressed using heuristics:

1. **Class Prediction Instability:** Class probabilities for a given object fluctuated across frames. To resolve this, we assigned the class with the highest cumulative probability across all frames in a clip.
2. **ID Inconsistency:** The tracker occasionally assigned different IDs to the same object across frames. We resolved this by merging IDs for objects of the same class that never co-occurred in the same frame. This sometimes resulted in temporary object disappearance ("not localized") when the object exited and re-entered the stall (e.g., a person leaving and returning during cleaning).

Table 2. Validation Metrics

| Model | Class | Precision | Recall | mAP50 | mAP50-95 |
|---|---|---|---|---|---|
| N | all | 0.966 | 0.922 | 0.96 | 0.843 |
| | *person* | *0.939* | *0.877* | *0.935* | *0.771* |
| | *horse* | *0.993* | *0.966* | *0.985* | *0.914* |
| S | all | 0.964 | 0.93 | 0.969 | 0.854 |
| | *person* | *0.94* | *0.889* | *0.947* | *0.785* |
| | *horse* | *0.987* | *0.97* | *0.991* | *0.922* |
| M | all | **0.975** | 0.916 | 0.97 | **0.87** |
| | *person* | *0.959* | *0.874* | *0.954* | *0.816* |
| | *horse* | *0.99* | *0.958* | *0.987* | *0.924* |
| L | all | 0.967 | **0.934** | **0.975** | 0.864 |
| | *person* | *0.962* | *0.895* | *0.957* | *0.806* |
| | *horse* | *0.972* | *0.972* | *0.992* | *0.921* |

## 2.4. From Multi-Object Tracking to Event Detection

To convert object trajectories into stall events (see Section 2.1), we followed a five-step procedure.

**Step 1 – Localization.** Each object in a frame is classified as either inside or outside the stall. This is determined by checking the intersection between the object's bounding box and the predefined floor polygon of the stall (see Figure 2). Objects that never enter the stall during a clip are discarded.

**Step 2 – Frame State Aggregation.** We aggregate the states of all detected objects of the same class (horse or person) into a single frame-level state using the rules in Table 3.

**Step 3 – Temporal Event Merging.** Consecutive frames with the same frame state are merged into temporal events. For instance, if frames k through n have the state "horse inside," a single event is created for that time range.

**Step 4 – Classifying 'Non-Localized' Events.** Events with non-localized objects are classified as either:
1. Inside (invisible): object is in a blind spot within the stall.
2. Outside (invisible): object has exited the stall.

This classification uses the last known frame of localization and evaluates the object's distance from the stall entrance and whether the bounding box touches the frame's edges.

**Step 5 – Inter-Clip Correction.** We adjust the classifications at clip boundaries. If the last event in clip *n–1* is "inside (invisible)" and the first event in clip *n* is "outside (invisible)," the latter is reclassified as "inside (invisible)," assuming the object remains in the blind spot.

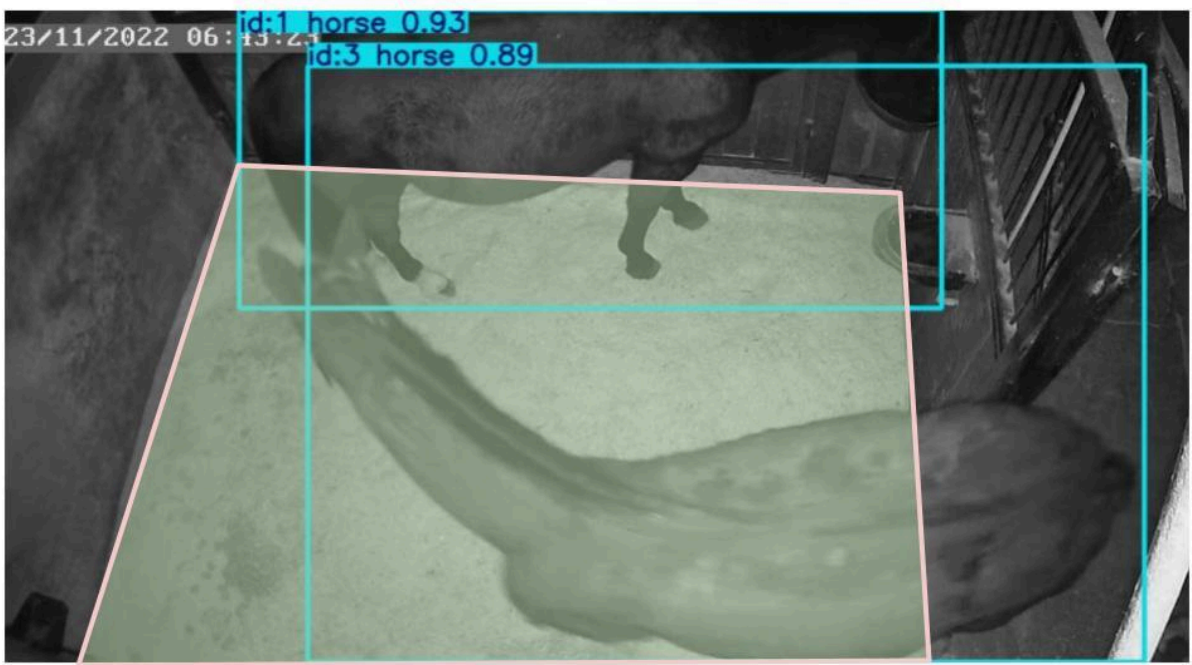

Figure 2. Floor polygon and bounding boxes of detected objects in the stall.

Table 3. Aggregation Rules for Frame State

| States of objects of the same class and located on the same frame | no object detected (empty frame) | - outside (N ≥ 1) | - not localized (N ≥ 1)<br>- outside (M ≥ 0) | - inside (N = 1)<br>- not localized (M ≥ 0)<br>- outside (K ≥ 0) | - inside (N ≥ 2)<br>- not localized (M ≥ 0)<br>- outside (K ≥ 0) |
|---|---|---|---|---|---|
| **Frame State** | outside (invisible) | outside (visible) | not localized | inside (visible) | multiple objects inside (visible) |

## 3. Results and Discussion

We evaluated the proposed monitoring system over a 24-hour period using video feeds from four cameras (k1, k2, k3, and k4). The system operated with a confidence threshold of 0.5, an IoU threshold of 0.5, and frame stride of 20.

During this period, some horses and people moved in and out of stalls. In total, the system processed 5,760 one-minute video clips. Among these, it generated non-empty event predictions for 360 clips:

1. 320 clips included only horse-related events,
2. 30 clips included only person-related events, and
3. 10 clips contained both horse and person events.

We selected the 10 clips containing both horse and person events as representative cases for qualitative analysis. For each clip, we compared the predicted events to manual annotations and interpreted the results (Table 4).

The analysis showed that the system accurately predicted horse-related events in all 10 clips. However, for person-related events, only 2 out of 10 clips had entirely correct predictions. The remaining clips exhibited various types of errors:

1. False positives: e.g., incorrect detection of "inside (invisible)" events.
2. Temporal shifts: misaligned start or end times of events.
3. False negatives: missed detections of people entering or exiting stalls.

The primary cause of these person-related errors was not sufficient YOLO's recall on the "person" class, consistent with the limited number of annotated person instances in the training set. Figure 3 illustrates a sequence of frames from the sixth of the analyzed clips, highlighting instances where YOLO failed to detect the presence of a person. These detection failures suggest that the model's performance could be substantially improved by augmenting the dataset with more labeled samples of people in various lighting and positional conditions.

Table 4. Qualitative analysis

| Video | Person Events (Predicted) | Person Events (GT) | Horse Events (Predicted) | Horse Events (GT) | Description (based on predicted events) | Description (based on GT) | Event error |
|---|---|---|---|---|---|---|---|
| 1 | 00, 47, outside_invisible<br>47, 48, outside<br>48, 54, outside_invisible<br>54, 57, inside<br>57, 60, inside_invisible | 00, 47, outside_invisible<br>47, 48, outside<br>48, 54, outside_invisible<br>54, 57, inside<br>57, 60, inside_invisible | 00, 48, outside_invisible<br>48, 60, inside | same | A person leads a horse into the stall | A person leads a horse into the stall | There is a time shift in the person's stall entry event. |
| 2 | 00, 58, inside_invisible<br>58, 60, inside | 00, 02, inside_invisible<br>02, 05, inside<br>05, 07, outside<br>07, 58, outside_invisible<br>58, 60, inside | 00, 60, inside | same | A person and a horse are inside the stall. | A person leaves the stall and then returns, while the horse remains inside. | The event of the person leaving the stall was missed. |
| 3 | 00, 03, inside<br>03, 60, inside_invisible | 00, 03, inside<br>03, 60, outside_invisible | 00, 60, inside | same | A person enters the stall while the horse is inside. | A person enters and exits the stall while the horse remains inside. | The event of the person leaving the stall was missed. |
| 4 | 00, 60, inside | same | 00, 60, inside | same | A person and a horse are inside the stall. | A person and a horse are inside the stall. | - |
| 5 | 00, 21, inside<br>21, 23, inside_invisible<br>23, 27, inside<br>27, 32, inside_invisible<br>32, 39, inside<br>39, 43, inside_invisible<br>43, 46, inside<br>46, 47, inside_invisible<br>47, 49, inside<br>49, 50, inside_invisible<br>50, 59, inside | 00, 60, inside | 00, 60, inside | same | A person and a horse are inside the stall. | A person and a horse are inside the stall. | The person did not enter the blind spot. |
| 6 | 00, 04, inside<br>04, 05, inside_invisible<br>05, 07, outside<br>07, 08, outside_invisible<br>08, 12, outside<br>12, 14, outside_invisible<br>14, 15, inside<br>15, 17, inside_invisible<br>17, 19, outside<br>19, 21, inside<br>21, 31, inside_invisible<br>31, 32, outside<br>32, 33, inside<br>33, 41, inside_invisible<br>41, 43, inside<br>43, 44, inside_invisible<br>44, 47, inside<br>47, 48, inside_invisible<br>48, 50, inside<br>50, 60, outside_invisible | 00, 55, inside<br>56, 60, inside_invisible | 00, 60, inside | same | A person exits and enters the stall multiple times while the horse remains inside. | A person and a horse are inside the stall. | False events of the person exiting and entering the stall. |
| 7 | 00, 19, outside_invisible<br>19, 21, inside<br>21, 22, inside_invisible<br>22, 38, inside<br>38, 60, outside_invisible | 00, 19, outside_invisible<br>19, 38, inside<br>38, 60, outside_invisible | 00, 60, inside | same | A person enters and exits the stall while the horse remains inside. | A person enters and exits the stall while the horse remains inside. | The person did not enter the blind spot. |
| 8 | 00, 14, outside_invisible<br>14, 15, outside<br>15, 36, outside_invisible<br>36, 38, outside<br>38, 40, outside_invisible<br>40, 41, inside<br>41, 60, inside_invisible | 00, 14, outside_invisible<br>14, 15, outside<br>15, 36, outside_invisible<br>36, 38, outside<br>38, 60, outside_invisible | 00, 40, inside<br>40, 60, outside_invisible | same | A person appears in the hall; The person enters the stall; The horse exits the stall. | A person appears in the hall; A person leads the horse out of the stall without entering it. | False event of person entering the stall. |
| 9 | 00, 15, outside_invisible<br>15, 16, outside<br>16, 17, inside<br>17, 24, inside_invisible<br>24, 26, inside<br>26, 27, inside_invisible<br>27, 28, inside<br>28, 31, inside_invisible<br>31, 32, outside<br>32, 60, outside_invisible | same | 00, 18, outside_invisible<br>18, 60, inside | same | A person leads the horse into the stall and then leaves. | A person leads the horse into the stall and then leaves. | The person did not enter the blind spot. |
| 10 | 00, 04, outside_invisible<br>04, 06, inside<br>06, 60, outside_invisible | 00, 02, outside_invisible<br>02, 06, inside<br>06, 60, outside_invisible | 00, 60, inside | same | A person enters and exits the stall while the horse remains inside. | A person enters and exits the stall while the horse remains inside. | There is a time shift in the person's stall entry event. |

Another error, observed in the sixth video, involved a person being detected outside the stall while actually remaining inside. As illustrated in Figure 4, the error occurred because the detected bounding box did not intersect with the stall's floor polygon—the lower part of the person's body was occluded by the horse. As a potential alternative for distinguishing between "inside" and "outside" positions, a separating line could be drawn at the base of the wall dividing the stall from the hallway. However, this method would only be applicable to stalls with similar camera angles and positioning, limiting its generalizability.

Regarding quantitative evaluation, computing event-based metrics such as t-IoU or mAP was deemed infeasible due to the high cost of manually annotating sufficiently large and temporally precise ground-truth labels. However, we recognize the importance of quantitative evaluation and plan to construct a manually labeled evaluation set in future work to support more rigorous benchmarking.

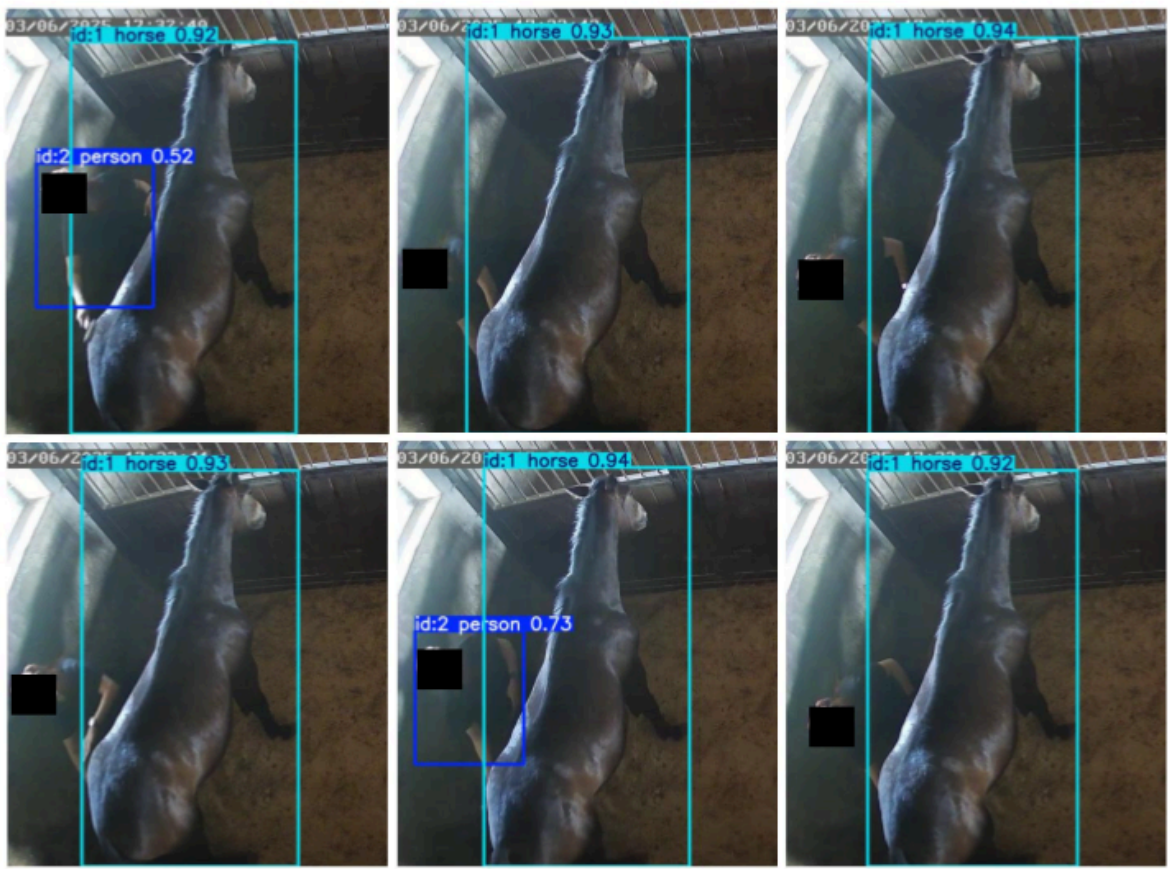

Figure 3. Examples of missed detections of a person across consecutive frames.

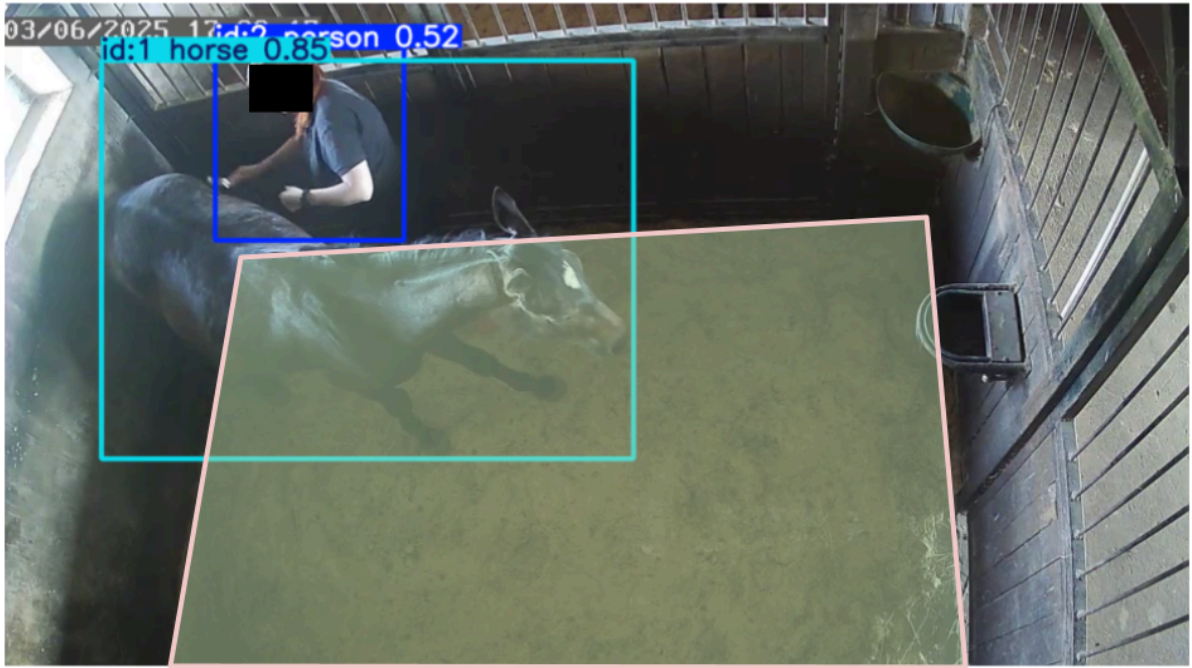

Figure 4. Example of an erroneous detection classifying a person as outside the stall.

## 4. Conclusion

In this study, we developed a prototype system for monitoring horses and people in stables using CV techniques. Our contributions are as follows:

1. **Dataset Development**: We collected and annotated a custom dataset of images containing horses and people, using CLIP and GroundingDINO to assist the labeling process.
2. **Model Training**: We trained a YOLOv11 model for robust detection of horses and people, achieving high accuracy for horses.
3. **Event Detection System**: We implemented a multi-stage system that tracks objects and detects temporal events—such as entry and exit from stalls—by analyzing bounding box trajectories.
4. **Qualitative Evaluation**: We conducted a qualitative assessment of the system's performance, which showed reliable detection of horse events but highlighted limitations in recognizing people due to data scarcity.

Although YOLOv11 demonstrated strong performance on the horse class (Precision = 0.99, Recall = 0.958), its performance on the person class was less reliable. This shortcoming was reflected in the final event detection output, particularly in cases involving human presence.

Future work will focus on expanding the training dataset, especially for the underrepresented "person" class, and on developing a quantitative evaluation dataset to support comprehensive, event-level benchmarking using metrics such as t-IoU and mAP.

Our system lays the groundwork for real-time behavioral monitoring in equine facilities, with the potential to support animal welfare, early anomaly detection, and improved management practices.

## Author Contributions

**Dmitrii Galimzianov**: Conceptualization, methodology, software development, formal analysis, investigation, visualization, and original draft preparation. **Viacheslav Vyshegorodtsev**: Resources, data curation, and review and editing of the manuscript. **Ivan Nezhivykh**: Data curation, software development. All authors have reviewed the results and approved the final version of the manuscript.

## Conflicts of Interest

The authors declare a financial interest in the product presented in this paper.

## Declaration of AI Use in Writing

During the preparation of this manuscript, the authors used OpenAI's ChatGPT to improve readability and language. After generating language suggestions, the authors reviewed, edited, and approved the content. They accept full responsibility for the final version of the manuscript.

## Ethical Statement

This research was conducted in accordance with the ethical guidelines and regulatory requirements of the Russian Federation, where the study took place. No experiments were conducted on human or non-human subjects that would require additional ethical approval.